\documentclass[lettersize,journal]{IEEEtran}
\usepackage{amsmath,amsfonts}
\usepackage{array}
\usepackage{subcaption}
\usepackage{textcomp}
\usepackage{stfloats}
\usepackage{url}
\usepackage{verbatim}
\usepackage{graphicx}
\usepackage{cite}
\usepackage{algorithm}
\usepackage{color}
\usepackage{algpseudocode}
\usepackage{booktabs}
\usepackage{adjustbox}
\usepackage{amsthm}

\begin{document}

%\title{Target to Source: Conditional Diffusion Model for Test-Time Adaptation}
\title{Two Simple Principles for Diffusion-Based Test-Time Adaptation}
\author{Kaiyu Song, Hanjiang Lai, Yan Pan, Kun Yue, Jian Yin}

\markboth{Journal of \LaTeX\ Class Files,~Vol.~14, No.~8, August~2021}%
{Shell \MakeLowercase{\textit{et al.}}: A Sample Article Using IEEEtran.cls for IEEE Journals}


\maketitle
\begin{abstract}
Recently, diffusion-based test-time adaptations (TTA) have shown great advances, which leverage a diffusion model
to map the images in the unknown test domain to the training domain. The unseen and diverse test domains make diffusion-based TTA an ill-posed problem. In this paper, we unravel two simple principles of the design tricks for diffusion-based methods.
Intuitively, \textit{Principle 1} says semantic similarity preserving. We should preserve the semantic similarity between the original and generated test images. \textit{Principle 2} suggests minimal modifications. This principle enables the diffusion to map the test images to the training domain with minimal modifications of the test images. In particular, following the two principles, we propose our simple yet effective principle-guided diffusion-based test-time adaptation method (PDDA).
Concretely, following Principle 1, we propose a semantic keeper, the method to preserve feature similarity, where the semantic keeper could filter the corruption introduced from the test domain, thus better preserving the semantics. Following Principle 2, we propose a modification keeper, where we introduce a regularization constraint into the generative process to minimize modifications to the test image. Meanwhile, there is a hidden conflict between the two principles. We further introduce the gradient-based view to unify the direction generated from two principles. Extensive experiments on CIFAR-10C, CIFAR-100C, ImageNet-W, and ImageNet-C with WideResNet-28-10, ResNet-50, Swin-T, and ConvNext-T demonstrate that PDDA significantly performs better than the complex state-of-the-art baselines. Specifically, PDDA achieves 2.4\% average accuracy improvements in ImageNet-C without any training process.
\end{abstract}
\section{Introduction}
\label{sec:intro}

\IEEEPARstart{D}{omain} shift problem~\cite{tcsvt,tcsvt-2,tcsvt-3} is the common challenge for computer vision, where the distribution of test data differs from that of the train data. For example, the nature corruption~\cite{BacktoSource} and the adversarial perturbation~\cite{adversarial} could cause the domain shift problem since they add additional noise to the test data.

Test-time adaptation (TTA)~\cite{TTT,Tent,TIPI,ITTT,kdd,MEMO} is one of the main paradigms against domain shift, which focuses on leveraging the unlabeled test data to alleviate domain shift problem at test time. According to the adjusting types of the test data, TTA could be classified into two categories: 1) The first category of TTA methods directly updates the training model (e.g., the classifier) using the unlabeled test samples~\cite{Tent}, and thus the training model can be adapted to the test distribution. And 2) \textit{Diffusion-based TTA methods}~\cite{BacktoSource, gda, sda}, which map the test samples back to the training distribution, and the training classifier can be directly used without retraining. The first method tries to fine-tune the training model unsupervised at test time. For example, TENT~\cite{Tent} is the first work to update the classifier by introducing the entropy loss. MEMO~\cite{MEMO} proposes the data augmentation method to enhance the entropy loss. ITTT~\cite{ITTT} further finds the test sample does not always provide useful information to optimize the entropy, thus proposing a better sample selection strategy. 

Instead of fine-tuning the training model to adapt to the test domain, diffusion-based TTA methods~\cite{BacktoSource} solve the domain shift problem by mapping the test data back to the training domain. These methods introduce a diffusion model that is pre-trained in the training domain.
Given the test sample from the unknown domain, the key is controlling the reverse process to generate the sample that the classifier can recognize correctly. 
DDA~\cite{BacktoSource} first proposes the diffusion-based TTA methods by leveraging the low-frequency filter as the metric to control the generation process. Then, GDA~\cite{gda} further introduces three complex metrics, i.e., classifier, CLIP~\cite{CLIP}, and UNet~\cite{SDE}, to enhance the controllability of the reverse process. SDA~\cite{sda} argues the gap between diffusion models and the classifier. It thus proposes fine-tuning the classifier again. 

In the view of the diffusion model, the diffusion-based TTA is inherently ill-posed since the test domain is unknown for the diffusion model. To illustrate this, we assume that the test domain contains the samples with an unknown level of Gaussian noise in the samples from the training domain. Suppose there are three different levels, such as 1) no Gaussian noise, 2) low-level Gaussian noise, and 3) high-level Gaussian noise. These three situations conflict. The diffusion model should do nothing for the first situation and needs more time steps to remove the noise in the third situation. The existing diffusion-based TTA methods will use the same reverse process to deal with three conditions, thus naturally needing the complex design to maintain high performance. This raises a question: \textit{are there common principles to guide the design of diffusion-based TTA?}

Motivated by this question, in this paper, we propose our simple yet effective common principles for diffusion-based TTA, that is, \textbf{preserving the semantics (P1) with as little modification as possible (P2)}.  
Concretely, we first propose two principles following the observation in previous works: \textbf{Principle 1} (P1): preserving the semantics. The goal of diffusion-based TTA is to ensure the classifier can recognize the generated samples correctly. Hence, the generated samples should have the same semantics as the original test samples. \textbf{Principle 2} (P2): minimal modifications. This principle ensures the diffusion model maps the test sample to the training domain with minimal modifications. The diffusion-based TTA methods depend on the posterior sampling~\cite{DPS} to guide the generation process. Following the posterior sampling, we use the two principles to formulate two conditional terms, i.e., two distance metrics. In this way, we propose our principle-guided diffusion-based test-time adaptation method (PDDA).

P1 suggests that we should try to maintain the semantics between the generated images and the test sample. To achieve it, we use patch-aware contrastive loss~\cite{freedom} as the distance metric,
The generated images should be closer to the reference image than other test images.
In our method, we can only access the pre-trained diffusion model at test time, which motivates us to choose the diffusion model as the feature extractor. There are many layers of the UNet in the diffusion model, and each layer contains different information~\cite{sda}. To preserve different semantics, we propose a patch-aware method to extract features from all layers into various groups and use the patch-aware contrastive loss to maintain the semantics of other groups.

P2 suggests that we should keep minor modifications of the test samples. The diffusion model trained on the training domain will gradually lead the test image to close the training domain. According to P2, we do not hope the generated image is too far from the test sample. To achieve this, we introduce mean square error (MSE) as the distance metric to reduce the modifications to the original test sample. Specifically, for time step $t$, we first obtain the posterior mean ~\cite{DPS}, which directly projects the $x_t$ back to the data distribution. Then, we minimize the square error between the posterior mean and the original test sample, requiring the generated $x_t$ to be closer to the original test sample.

There may be a hidden conflict between the two principles: one needs us to edit the samples, and the other limits us to minor modifications. To alleviate this, we introduce a gradient projection method to unify two gradients' directions. Therefore, our PDDA could perform better using only two simple distance metrics. Following the prior works with different pre-trained classifiers, experimental results on CIFAR-10C, CIFAR-100C, ImageNet-C, and ImageNet-W prove that PDDA achieves the SOTA performance. 

To sum up, the main contributions of this paper are:
\begin{itemize}
    \item We proposed PDDA, a diffusion-based TTA method with two simple metrics to map the test images from the test domain to the training domain.
    \item We show the potential of the diffusion-based TTA method, where the simple design could also achieve high performance.
    \item The experimental results show that PDDA achieves state-of-the-art results via various diffusion-based methods.
\end{itemize}

\section{Related Work}
\label{sec:related work}
%pass-view check
\textbf{Test-time adaptation.}
Domain shift problem exists in various computer vision tasks~\cite{tcsvt,tcsvt-2,tcsvt-3}. For example, in the image classification~\cite{tkde-domain_adaption,tcsvt-5,tcsvt-6} task, domain shift emerges when there is an unknown type of noise in the images of the test domain.
In the segmentation tasks~\cite{tcsvt-4} task, domain shift emerges when the trend in a certain interval has changed significantly, such as the price trend fluctuation in the shares market. In the entity resolution~\cite{vldb-domain_adaptation} task, the domain shift exists when the data distribution of the target ER dataset is different from the source ER datasets. 

TTA is one of the settings for domain shift problems, where TTA assumes that we cannot access the training data. In other words, we can only access the test data in test time without its label. Therefore, TTA focuses on introducing the unsupervised manner~\cite{tkde-domain_adaption_Unsupervised,icde-domian_adaptation1}. TTA methods could be further categorized into training-based adaptation, directly updating the training model and diffusion-based adaptation.

The main challenge for the training-based adaptation methods is updating the training model correctly without labels against unknown domain shift~\cite{Tent}. For example, the test-time train~\cite{TTT} has been proposed by creating an auxiliary task to automatically label the target data, e.g., predicting the rotation angle. This task leveraged a new rotation angle label to replace the original class label and train the model based on the collected test data. 

Then, TENT~\cite{Tent,ITTT} proposed the first unsupervised manner for the TTA by entropy minimization. TENT was based on an observation that images with small entropy always have a high likelihood of being classified correctly versus a low likelihood. Meanwhile, entropy is a label-free metric~\cite{Tent} and thus could be used to train the model unsupervised. Meanwhile, Yang \textit{et al.}~\cite{sigmod-domian_adpation} proposed the new principle to guide and unify the view of training-time and test-time learning, which increased the robustness of the training model, thus defending against the domain shifts. Yang \textit{et al.}~\cite{sigmod-test_time} improved the entropy method by introducing calibrated entropy minimization.

\textbf{Diffusion-based adaptation.} Diffusion models~\cite{tkde-diffusion_survery} show the potential to solve various downstream tasks, such as improving robust generalization~\cite{sigmod-diffusion} and dealing with tabular data~\cite{icde-diffusion}.  Diffusion-based adaptation tries to leverage the prior knowledge of the diffusion model pre-trained in the training domain to adjust the test inputs for TTA. 

For example, Diffpure~\cite{DiffurePure} first leveraged the unconditional diffusion model to alleviate the domain shift caused by the adversarial perturbation. Diffpure used the forward process to cover the noise in the test-time, where the test-time data will add unknown types of adversarial perturbation to attack the DNNs. Then, the reverse process will remove the Gaussian noise with the adversarial perturbation. This reveals a natural property that the reverse process of the diffusion model could let the images be close to the training domain.

Following this, Gao \textit{et al.}~\cite{BacktoSource} introduced the posterior sampling method~\cite{DPS} to solve the domain shift caused by natural corruption, where the sample in the test domain is used as the condition to guide the generation process. Hu \textit{et al.}~\cite{skin_diffusion} trained a diffusion model to solve the skin lesion classification based on TTA. Shim \textit{et al.}~\cite{3dpoint_diffusion} leveraged the diffusion model to solve the 3D could point-based TTA. Then, in the image classification tasks, following the DDA, GDA~\cite{gda} added additional guidance terms based on the vision-language models, such as CLIP~\cite{CLIP}, to offer more prior knowledge into the reverse process. Thus, the sample could maintain a high quality while being closer to the training domain. 

Interestingly, Guo \textit{et al.}~\cite{sda} further proposed a new setting in which we could train the classifier again in the unified domain, i.e., the generation domain of the diffusion model. This is motivated by the observation that even though the diffusion model was pre-trained in the training domain, there is a natural gap between the diffusion model and classifier, thus leading the domain to shift again. To solve this problem, Guo \textit{et al.} further proposed a complex framework to train the classifier in the generated images from the diffusion model by combining an unconditional diffusion model and a conditional diffusion model.

The framework of the diffusion-based adaptation starts from a single unconditional diffusion model and then uses two diffusion models, which tend to become complex. However, there are various common points among these designs. This motivates us to explore common principles to enable the framework's design effectively.

\section{Preliminary}
\label{sec:preliminary}

\textbf{Problem formation.} 
Following previous settings~\cite{BacktoSource,gda}, we define the pre-trained classifier as $p_{\phi}(y|x)$ and the pre-trained diffusion model as $s_{\theta}$. Then, we define the sample sampling from the test domain as $x^{test}$. We can only access the $x^{test}$, $p_{\phi}(y|x)$, and $s_{\theta}$ in the test time. Meanwhile, the domain of $x^{test}$ is unknown.

\textbf{Diffusion-based adaptation.} Given the inputs $x^{test}$, we use the forward process~\cite{DPS} to map it to the Gaussian noise:
\begin{equation}
    x_t = \sqrt{\hat{\alpha}_{t}}x^{test} + \sqrt{1-\hat{\alpha}_{t}}\epsilon,
    \label{eq:forward process}
\end{equation}
where $\hat{\alpha_{t}}$ is the noise schedule related to the time step $t$ and $\epsilon \sim \mathcal{N}(0,1)$ is the Gaussian noise.

Starting from a large time step $t = T$, $x^{t}$ will become the pure Gaussian nose. Then, by the reverse process~\cite{DPS}:
\begin{equation}
    x_{t-1} = \frac{1}{\sqrt{1-\beta_{t}}}(x_{t} - \frac{\beta_{t}}{\sqrt{1-\sigma(t)}}s_{\theta}(x_{t},t)) + \phi_{t}*\epsilon,
    \label{eq:reverse process}
\end{equation}
where $\sigma(\ast)$, $\phi_{t}$, and $\beta_{t}$ are the constant from the noise schedule related to the time step $t$.

Given the diffusion model $s_{\theta}(x_{t},t)$ and the test sample $x^{test}$, we incorporate the posterior sampling for the diffusion model to guide the generation process. Concretely, we use $x^{test}$ as the conditional guidance in the diffusion model $s_{\theta}(x_{t},t)$. Please note that the score function of $s_{\theta}(x_{t},t)$ is the gradient of the log probability $\nabla_{x_{t}}\log p(x_{t})$. Since the pre-trained diffusion model is an unconditional generation model, the test sample $x^{test}$ as the conditional guidance in the diffusion model can be formulated as $\nabla_{x_{t}}\log p(x_{t}|x^{test})$. By the Bayesian rule, we have:
\begin{equation}
    \nabla_{x_{t}}\log p(x_{t}|x^{test}) =  \underbrace{\nabla_{x_{t}}\log p(x_{t})}_{\text{Unconditional Term}} +   \underbrace{\nabla_{x_{t}}\log p(x^{test}|x_{t})}_{\text{Conditional Term}}.
    \label{eq:dps}
\end{equation}

In this way, the conditional term tries to keep the label semantic and gradually let the generated image return to the training domain, i.e., the source domain, during the reverse process.

The next step is to estimate the conditional term. By the DPS~\cite{DPS}, we could first leverage the MMSE estimation to map the $x_{t}$ from the noise domain to the data domain:
\begin{equation}
    \hat{x}_{0|t} = \frac{1}{\sqrt{\hat{a}_{t}}}(x_{t} + (1-\hat{\alpha}_{t})*s_{\theta}(x_{t},t)).
    \label{eq:mmse}
\end{equation}
Then, the conditional term could be changed to $\nabla_{x_{t}}\log p(x^{test}|x_{t})\approx \nabla_{x_{t}}\log p(x^{test}|\hat{x}_{0|t})$. By defining distance metric $D(x^{test},\hat{x}_{0|t})$ to measure the distance between $x^{test}$ and $\hat{x}_{0|t}$, we could guide the diffusion model to generate the images closer to the $x^{test}$ as:
\begin{equation}
    \nabla_{x_{t}}\log p(x^{test}|\hat{x}_{0|t}) \approx -R\nabla_{x_{t}} D(\hat{x}_{0|t}, x^{test}),
    \label{eq:conditionterm}
\end{equation}
where $R$ is the scale factor to control the conditional term, which could be calculated directly based on DPS~\cite{DPS}.

Eq.~\ref{eq:conditionterm} is the case that only uses one metric to guide the generation process, which could be extended to the multiple distance metrics~\cite{freedom} used in GDA as:
\begin{equation}
    \nabla_{x_{t}}\log p(x^{test}|x_{t}) = \sum_{i}^{n}{-R_{i}\nabla_{x_{t}} D^{i}(\hat{x}_{0|t}, x^{test})},
    \label{eq:multi-conditionalterm}
\end{equation}
where $R_{i}$ is the scale factor related to the $i$-th metric $D^{i}$ and $n$ is the number of the metrics.

\textbf{Gradient magnitude.} Given two gradients $v_{1}$ and $v_{2}$, following the gradient magnitude similarity~\cite{pgd_gradient}, the conflict between $v_{1}$ and $v_{2}$ could be measured as:
\begin{equation}
    \Phi(v_{1},v_{2}) =  \frac{2||v_{1}||_{2}||v_{2}||_{2}}{||v_{1}||_{2}^{2} + ||v_{2}||_{2}^{2}}.
    \label{eq:conflict}
\end{equation}
When the magnitude of two gradients is the same, Eq.~\ref{eq:conflict} is 1, as the two gradients are in totally different directions, Eq.~\ref{eq:conflict} tends to be 0.

\textbf{Motivation.} Eq.~\ref{eq:conditionterm} shows the conditional term leverages the gradient of the metrics as the directions to guide the generated $x_{t}$ to keep the semantic with the test sample. The unconditional term is the constraint to ensure $x_{t}$ will close to the training domain. Intuitively, the principle exists to help us design the conditional term such that the generated images could be close to the training domain while keeping the semantics, which motivates us to explore these principles instead of letting the conditional term always be complex.

\section{Method}
\label{sec:methodology}

\begin{figure*}
    \centering
    \includegraphics[width=0.9\linewidth]{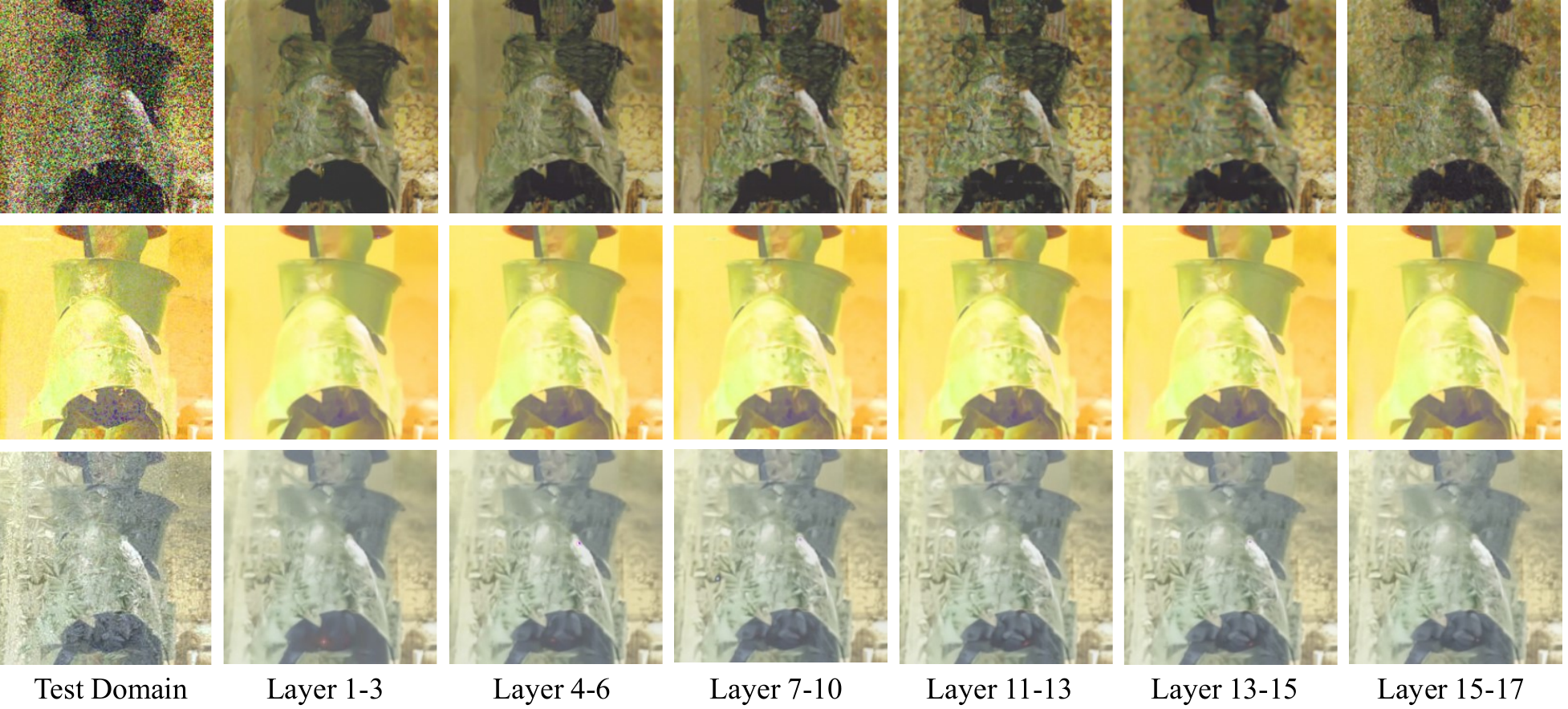}
    \caption{An illustration of the different information among features of different layers in UNet. We visualize the feature extracted from all layers by using features of a single layer to calculate $f^1$. It can be noticed that the details of the generated images will have little differences to prove that there is different information. This is obvious in the noise test domain since deep layers will contain the extra noise with more details.}
    \label{fig:inner_conflict}
\end{figure*}

\begin{figure*}[h!]
    \centering
    \includegraphics[width=\textwidth]{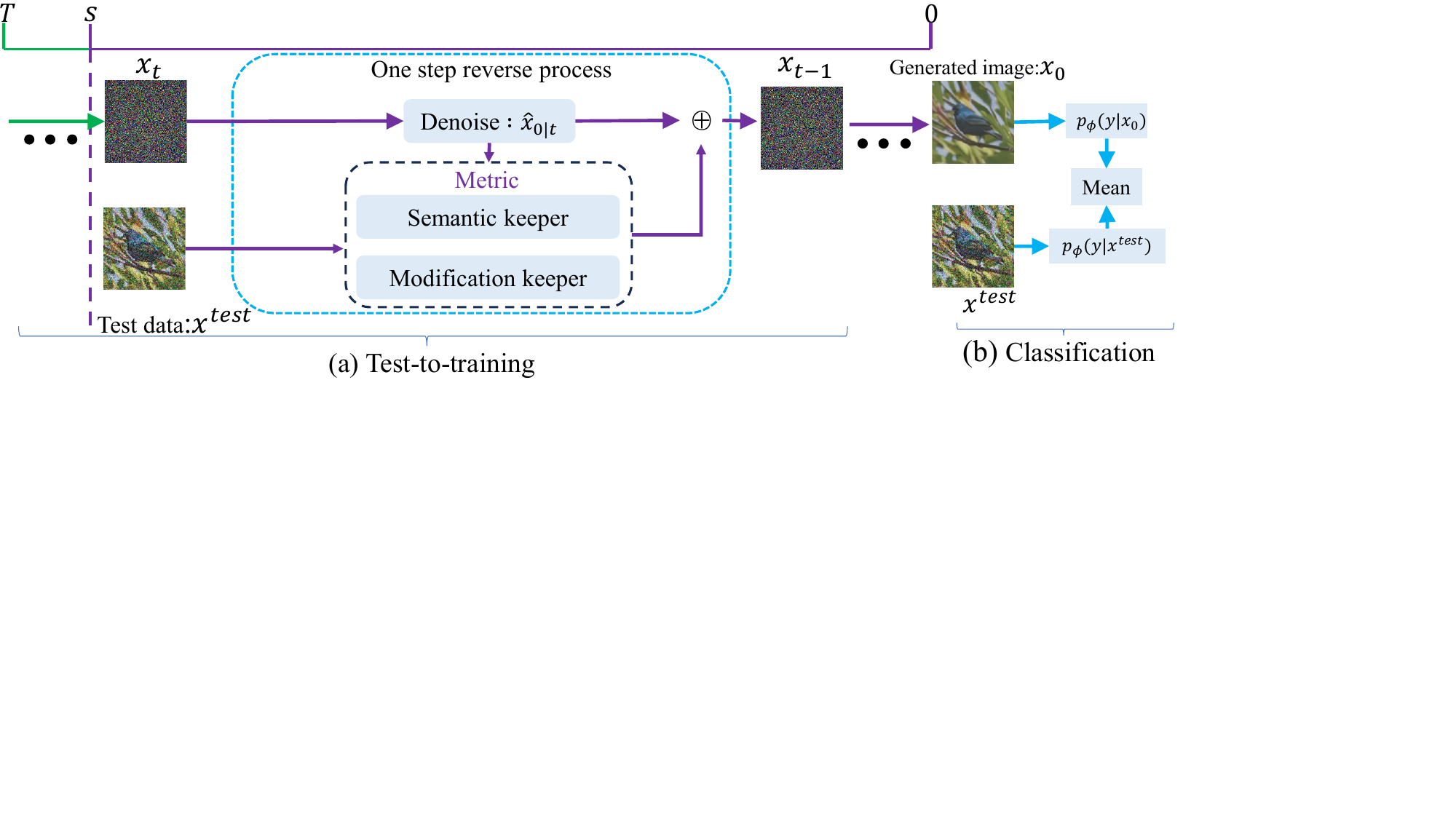}
    \caption{An overview of the proposed PDDA, where denoise is to estimate the $\hat{x}_{0|t}$. We implement the conditional term in the interval $t \in [s,0)$, and the interval $t\in [T,s)$ is the reverse process without the conditional term. $x_{t}\rightarrow x_{t-1}$ represents the one step of the reverse process. We first achieve the process that maps the image from the test domain to the training domain (a) to generate the $x_{0}$ and use both $x_{0}$ and $x^{test}$ to finish the classification (b).} 
    \label{fig:SEDA}
\end{figure*}

First, we indicate two principles of diffusion-based adaptations. Then, we propose the principle-guided diffusion-based test-time adaptation method (PDDA). PDDA contains a semantic keeper $f^{1}$ to preserve the semantic similarity by P1 and a modification keeper $f^{2}$ as the constraint to control the modification by P2. By leveraging two keepers, PDDA could calculate the conditional term of Eq.~\ref{eq:multi-conditionalterm}. The overall framework for PDDA is shown in Fig.~\ref{fig:SEDA}.

\textbf{Principle 1}. The target of the TTA is to let the classifier get the correct label for $x^{test}$. This requires the generated images $x_{0}$ to keep the same label semantic with the $x^{test}$ during the generating process. Thus, preserving semantics is vital for diffusion-based TTA methods (i.e., Principle 1 in this paper).

This requires us to challenge the gap of the domain shift between the $x^{test}$ and samples in the training domain related to $x^{test}$. For example, given a photo of the dog, one is clean, and the other is with Gaussian noise. We should try to keep the main object, i.e., the dog, unchangeable while trying to ignore the Gaussian noise. Thus, we propose the semantic keeper. Concretely, the semantic keeper is motivated by contrastive learning since our target is similar to it. For example, we should push the patch that includes the dog pixels into the same group and then pull out the patch without the dog pixels. Then, the correlation among patches may help us weaken the influence of the Gaussian noise, thus preserving the semantics as much as possible. Thus, the semantic keeper introduces the patch-aware contrastive loss~\cite{sda} as the distance metric to ensure P1.

The patch-aware contrastive loss needs a feature extractor~\cite{sda} to get the feature patch. It requires the features to be good enough to represent the images. In our method, we only access the pre-trained diffusion model and the classifier. Thus, the pre-trained diffusion model is a good choice for extracting the feature. We leverage the UNet, the backbone of the diffusion model, as the feature extractor. There are many layers under the UNet, and different layers offer different information. To illustrate this, as shown in Fig.~~\ref{fig:inner_conflict}, we conduct the visualization to show how such information influences the final generated results. The conclusion is that the different layers contain different information. To avoid falling into a complex selection strategy, the features from all layers should be used, which requires calculating the patch-aware contrastive loss at various times. This motivates us to propose a patch-aware method to extract features from all layers while calculating patch-aware contrastive loss only once. 

The idea is simple: we could first aggregate all features into one without losing the semantics. Thus, we only need to calculate patch-aware contrastive loss once. At the reverse time step $t^{\ast}$, we use the UNet of $s_{\theta}(\ast,t^{\ast})$ as the encoder to extract the feature map for the generated image $x_{t}$ and $x^{test}$. Then, we define $E \in \mathbb{R}^{C \times H \times W}$ as the output of UNet so that $E = f_e(\ast) $, $C$ is the channel and $(H, W)$ is the resolution of the feature maps. Suppose the number of layers is $L$; then, we can obtain $L$ feature maps with different channels, heights, and widths. According to the channels, we split the $L$ feature maps into $M$ groups: $\{G_{1}, G_{2}, \cdots, G_{M}\}$, where each group has the same number of channels, and $G_{m}$ is the number of layers in the $m$-th group with $\sum_{m=1}^M  G_{m} = L$. 

For the $G_{m}$ group, the number of channels for all feature maps is the same, denoted as $C_m$.
We denote the minimum resolution as $(H_{m}^{\min}, W_{m}^{\min})$. We reshape all the feature maps in the $G_{m}$ group into $E^{'}_{m} \in \mathbb{R}^{C_{m} \times H_{m}^{\min} \times W_{m}^{\min}}$. 

After reshaping, all the feature maps in the same group have the same channel and resolution. We finally use the sum operator to sum the feature maps in the same group followed by a batch norm layer, denoted as $\hat{E}_{m}$, where $\hat{E}_{m} = \text{BatchNorm}(\sum_{E \in E^{'}_{m}}{E})$. 

We use the sum operator to capture the correlation in each group since the feature with high correlation frequently appears in feature maps extracted from different layers. Thus, the sum operator can intensify their signal. The batch norm layer is used to smooth the gradient. The aggregated feature maps are the $\hat{E}_{m}$.

In the end, similar to the ViT~\cite{VIT} which divides the image into $P \times P$ patches, we reshape the $ \hat{E}_{m}\in \mathbb{R}^{C_{m}\times H_{m}^{min} \times W_{m}^{min}}$ into 2D sequence patches $e^{p}_{m}$~\cite{VIT}.
Based on the feature patches extracted from $x^{test}$ and $\hat{x}_{0|t}$, we build the pairs: $\{ (e^{p,1}_{1}, e^{p,2}_{1}),...,(e^{p,1}_{m}, e^{p,2}_{m})\}$, where $e^{p,1}_{m}$ denotes the patch from $x^{test}$ and $e^{p,2}_{m}$ denotes the patch from $\hat{x}_{0|t}$ respectively. For each pair, we calculate the patch-aware contrastive loss:
\begin{equation}
             \ell_{cl}(e^{p,1}_{m},e^{p,2}_{m}) = \frac{\exp(s_{i,i}/\tau)}{\sum_{k\neq i}\exp(s_{i,k}/\tau) + \exp(s_{i,i}/\tau)},
\end{equation}
where $s_{i,i}$ is the patch-aware similarity matrix related to pair $(e^{p,1}_{m},e^{p,2}_{m})$.

Therefore, the semantic keeper could be finally calculated based on all feature patches as follows:
\begin{equation}
    f^1(\hat{x}_{0|t}, x^{test}) = \nabla_{x_{t}}\frac{1}{M}\sum_{i=1}^{M}\ell_{cl}(e^{p,1}_{i},e^{p,2}_{i}).
    \label{eq:closs}
\end{equation}

\begin{figure}
 \centering
        \includegraphics[width=\linewidth]{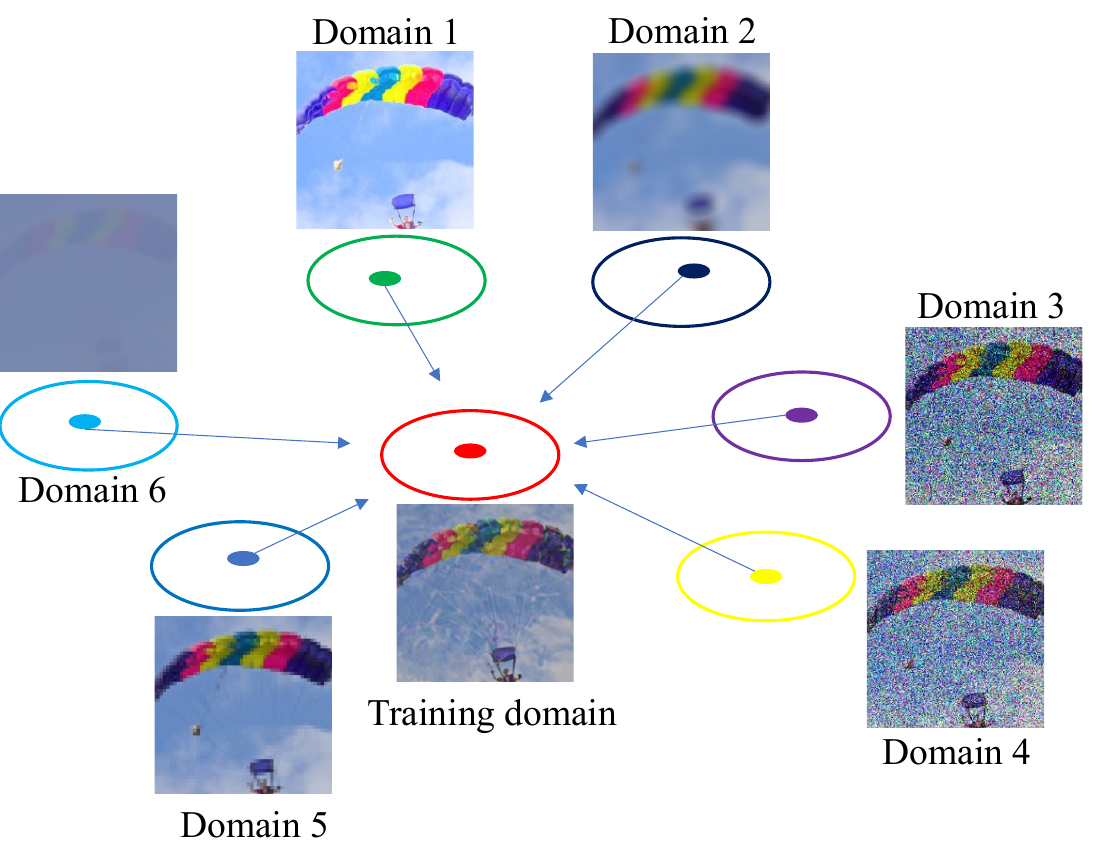}
        \caption{An illustration of the ill-posed problem made in different domains. We use different cycles with different colors to represent different domains. The \textcolor{red}{red cycle} represents the training domain. The diffusion-based methods need to narrow the distance between the $x^{test}$ and $x^{src}$, which needs to let random samples from unknown different domains close to the training domain.}
        \label{fig:ill-posed}
\end{figure}

\textbf{Principle 2.} P2 derives from the ill-posed problem. Diffusion-based adaptation needs to narrow the distance between the $x^{test}$ and samples in the training domain. The $x^{test}$ is from the unknown domain with infinity combinations. To illustrate this, we give an example in Fig.~\ref{fig:ill-posed}. To ensure it works for most conditions, we have to rely on the prior knowledge of the unconditional term to return to the test domain. Thus, a minor modification is a sub-optimal solution we choose. Minor modification will at least make sure the generated images will not deviate from the data distribution and be slightly close to the $x^{src}$. Thus, we propose a constraint with the MSE distance as the modification keeper to limit the pixel change, which could be formulated as:
\begin{equation}
    f^2(\hat{x}_{0|t},x^{test}) = \nabla_{x_{t}} ||x^{test}-\hat{x}_{0|t}||_{2}.
    \label{eq:mse}
\end{equation}

\begin{figure}
 \centering
     \begin{subfigure}[b]{0.22\textwidth}
         \centering
         \includegraphics[width=\linewidth]{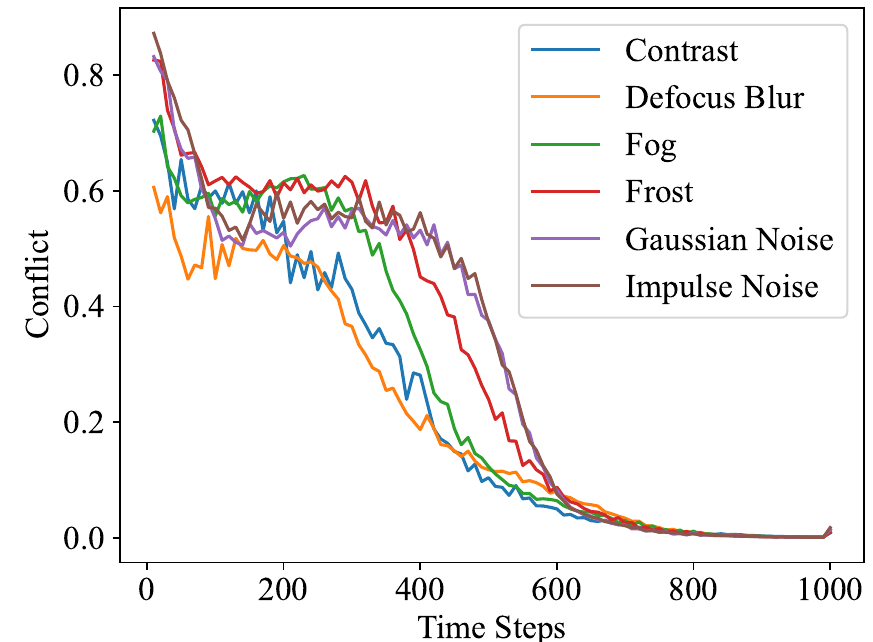}
         \caption{W/o the projection.}
     \end{subfigure}
     \begin{subfigure}[b]{0.22\textwidth}
         \centering
         \includegraphics[width=\linewidth]{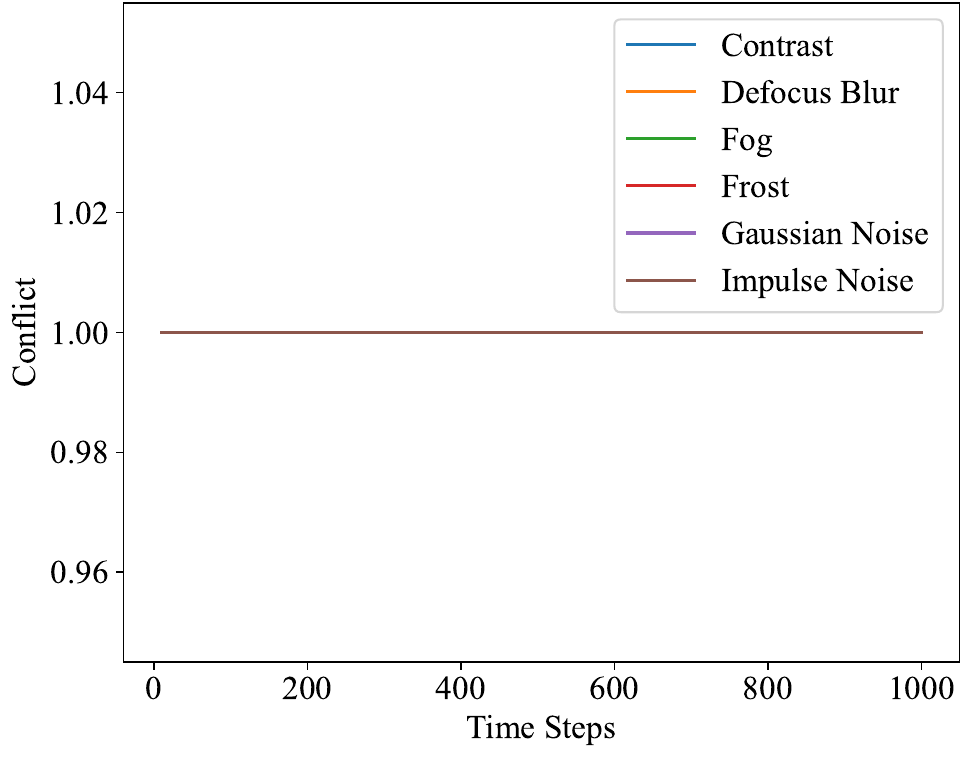}
         \caption{W/ the projection.}
     \end{subfigure}
        \caption{An illustration of the conflict in two guidance. w/ means using the gradient projection, and w/o means not using the gradient projection. We visualize the gradient magnitude similarity in Eq.~\ref{eq:conflict} based on samples from different test domains. It can be noticed that gradient magnitude similarity tends to zero without the gradient projection shown in (a). Then, our method can significantly increase the similarity.}
        \label{fig:outer_conflict}
\end{figure}

\textbf{Gradient projection.} Based on semantic and modification keepers, we could calculate the conditional terms as follows:
\begin{equation}
\begin{split}
    &\nabla_{x_{t}}\log p(x^{test}|x_{t}) = \nabla_{x_{t}}\log p(f^{1},f^{2}|x_{t})\\
    &= -R(f^{1}(\hat{x}_{0|t},x^{test}) + f^{2}(\hat{x}_{0|t},x^{test})).
\end{split}
\label{eq:multi_guidance}
\end{equation}
Eq.~\ref{eq:multi_guidance} shows that there may be a conflict between $f^{1}$ and $f^2$ since $f^1$ have to edit the images and $f^2$ have to ensure the minor modification. To illustrate this, as shown in Fig.~\ref{fig:outer_conflict}, we visualize the conflict between $f^{1}$ and $f^2$. This is similar to the gradient descent for multiple loss functions, where the conflict among loss functions causes the gradient to be unstable and further leads to falling into the sub-optimal solution. This conflict may lead to a failed generation~\cite{pgd_gradient}. Following the gradient projection~\cite{pgd_gradient}, we first normalize $f^{1}$ and $f^2$ as:
\begin{equation}
\begin{split}
        g^{1} &= \frac{f^{1}(\hat{x}_{0|t},x^{test})}{||f^{1}(\hat{x}_{0|t},x^{test})||_{2}}\\
        g^{2} &= \frac{f^{2}(\hat{x}_{0|t},x^{test})}{||f^{2}(\hat{x}_{0|t},x^{test})||_{2}}.
\end{split}
\end{equation}
Then, we choose $f^1$, the semantic keeper, as the main direction since keeping the semantic is the key for TTA. Thus, we project the $f^2$ onto the normal plane of the $f^1$ as:
\begin{equation}
f^{2}(\hat{x}_{0|t},x^{test}) = g^{2} - \frac{g^{2}g^{1}}{||g^{1}||_{2}}g^{1}.
    \label{eq:projection_g2_to_g1}
\end{equation}
In this way, we alleviate the conflict between the semantic and modification keepers. 

\textbf{Sampling and ensemble strategy.} In the early phase, $\hat{x}_{0|t}$ is close to the pure Gaussian noise, which will lead to much bias for Eq.~\ref{eq:dps}. Meanwhile, the acceleration method such as DPM-Solver++\cite{dpm} is sensitive to the bias. To avoid these problems and let PDDA be implemented in accelerating methods following the empirical finding~\cite{freedom}, it is better to implement the conditional term in the latter phase, defined as the interval $[s,0)$. Therefore, following the FreeDom~\cite{freedom}, we set $s=50\%T$.

Meanwhile, to leverage the prior knowledge of the classifier from the test domain further, we use a simple ensemble strategy:
\begin{equation}
    y^{pred} = 0.5*(p_{\phi}(y|x_{0}) + p_{\phi}(y|x^{test})),
    \label{eq:ensemble}
\end{equation}
where $y^{pred}$ is the final prediction of the label by the classifier.

Algorithm 1 summarizes the entire method. In this way, with two principles, PDDA achieves a simple design with semantic and modification keepers. The semantic keeper contains a contrastive loss as the distance metric with a simple feature extractor method, and the modification keeper only contains an MSE metric. Based on our simple design, PDDA could successfully generate images that are close to the training domain.
\begin{algorithm}[t]
    \caption{The overall algorithm for PDDA} \label{al:stage1}
    \begin{algorithmic}[1]
     \Statex \textbf{Input:} $x^{test}$, $T$, $\epsilon_{\theta}$, $\sigma(\cdot)$, $p_{\phi}(y|x)$, and $t^{\ast}$
     \Statex \textbf{Output:} $y^{pred}$
    \State Calculate $x_{T}$ by Eq.~\ref{eq:forward process}
    \For{$t$ in $[T,T-1,...,1]$} 
        \State Calculate $\hat{x}_{t}$ by Eq.~\ref{eq:dps} and $x_{t-1}$ by Eq.~\ref{eq:reverse process}
        \If{$t \in [50\%T,0)$}
        \State Calculate $f^{1}$ by Eq.~\ref{eq:closs}
        \State Calculate $f^{2}$ by Eq.~\ref{eq:projection_g2_to_g1}
        \State Calculate conditional term by Eq.~\ref{eq:multi_guidance}
        \State Calculate $x_{t-1}$ by Eq.~\ref{eq:dps}
            \Else
                \State $x_{t-1} \gets x_{t-1}$
        \EndIf
    \EndFor \label{ref:end}
    \Statex Calculate $y^{pred}$ by Eq.~\ref{eq:ensemble}
    \Statex\textbf{Return:} $y^{pred}$
    \end{algorithmic}
\end{algorithm}

\section{Experiment}
\label{sec:experiment}
\subsection{Experiment Settings}

\textbf{Datasets and network architectures}. We consider four datasets for evaluation: CIFAR-10C, CIFAR-100C, ImageNet-W~\cite{imagenetw}, and ImageNet-C~\cite{CIFAR10_cifar100c}. Meanwhile, we compare state-of-the-art diffusion-based methods such as DDA~\cite{BacktoSource} and DiffPure~\cite{DiffurePure}, GDA, and SDA~\cite{sda}. To show the improvement of our method, we also compare various state-of-the-art training-based methods such as TENT~\cite{Tent} and MEMO~\cite{MEMO}. For pre-trained classifiers, we consider two widely used backbones for CIFAR-10C and CIFAR-100C: WideResNet-28-10~\cite{WideResNet}. For ImageNet-C and ImageNet-W, we consider the ResNet-50~\cite{Resnet26}, ConvNext-T~\cite{Convnext}, and Swin-T~\cite{Swin} as the backbones.

\begin{table*}
    \centering
    \caption{Classification accuracy (\%) against the different types of corruption on CIFAR-10C under WideResNet-28-10, where Gaussian-N is the Gaussian noise.}
    \begin{adjustbox}{width={\textwidth},totalheight={\textheight},keepaspectratio}
    \setlength{\tabcolsep}{2px}{
    \begin{tabular}{c c c c c c c c c c c c c c c c c c c c c}
        \toprule
        \textbf{Method} & \scriptsize  \textbf{Brightness} & \scriptsize \textbf{Contrast} & \scriptsize \textbf{Defocus} & \scriptsize \textbf{Elastic} & \scriptsize \textbf{Fog}& \scriptsize \textbf{Frost}& \scriptsize\textbf{Gaussian-B}& \scriptsize \textbf{Gaussian-N}&\scriptsize \textbf{Glass}& \scriptsize \textbf{Impulse}& \scriptsize \textbf{JPEG}& \scriptsize \textbf{Motion}& \scriptsize \textbf{Pixel}& \scriptsize \textbf{Saturate}& \scriptsize \textbf{Shot}&\scriptsize \textbf{Snow}&\scriptsize \textbf{Spatter}&\scriptsize \textbf{Speckle}&\scriptsize \textbf{Zoom}&\scriptsize \textbf{Avg.}  \\
        \midrule
        \textbf{Training-based}\\
        \midrule
        MEMO~\cite{MEMO} & 93.5& 90.7& 88.95& 81.25& \textbf{90.74}& 85.6& 85.04& 72.43& 60.04& 61.27& 65.4& 74.0& 68.42& 73.44& 55.25& 57.14& 57.59& 48.88& 54.02& 71.78 \\
        TENT~\cite{Tent} &\textbf{95.05}& \textbf{88.8}& \textbf{90.89}& 81.51& 86.98& 85.68& \textbf{85.94}& 74.74& 69.01& 73.7& 77.34& 80.73& \textbf{84.11}& 90.1& 78.91& 82.29& 80.73& 79.69& \textbf{85.42}& 82.72 \\
        \midrule
        \textbf{Diffusion-based} \\
        \midrule
        DiffPure~\cite{DiffurePure} &  78.96& 68.96& 65.21& 69.79& 78.96& 57.92& 57.29& 48.96& 41.04& 50.21& 60.42& 61.04& 65.0& 78.75& 51.46& 65.21& 68.33& 56.04& 62.08& 62.4\\
        DDA~\cite{BacktoSource}& 83.63& 76.18& 76.47& 72.55& 80.29& 64.8& 63.63& 55.49& 53.24& 55.39& 68.73& 67.55& 68.04& 81.08& 63.04& 69.22& 75.59& 60.0& 68.04& 68.58 \\
        PDDA (Our) & 92.35& 80.0& 83.92& \textbf{86.47}& 87.06& \textbf{89.22}& 78.24& \textbf{82.75}& \textbf{72.25}& \textbf{78.04}& \textbf{84.31}& \textbf{82.45}& 82.45 & \textbf{92.45}& \textbf{83.33}& \textbf{85.88}& \textbf{90.39}& \textbf{85.39}& 80.49& \textbf{84.08} \\ 
        \bottomrule
    \end{tabular}}
    \end{adjustbox}
    \label{tab:cifar10c-wideres}
\end{table*}

\textbf{Diffusion model settings}. We use the unconditional CIFAR-10 checkpoint of EDM offered by NVIDIA \cite{edm} for our method on CIFAR-10C datasets. We fine-tune the unconditional CIFAR-10 checkpoint based on CIFAR-100 for CIFAR-100C following the training method offered by NVIDIA \cite{edm}. For ImageNet-C, we use the pre-trained diffusion model offered by Dhariwal \textit{et al.}~\cite{diffusionbeatgan}. We evaluate our model on a single RTX4090 GPU with 24GB memory. To prove the proposed guidance method could work with the accelerating sampling method for the diffusion model, we directly implement our method based on the DPM-Solver++~\cite{dpm}. For all experiments, we use $T=100$.

\textbf{Evaluation metrics}. Similar to the Gao \textit{et al.}~\cite{BacktoSource}, we evaluate our method based on the accuracy under nineteen types of corruptions~\cite{CIFAR10_cifar100c} in  CIFAR-10C and  CIFAR-100C and 15 types of corruptions on ImageNet-C. We also evaluate our method based on the accuracy under handcraft work corruptions in ImageNet-W. Meanwhile, to reduce the computation cost, we evaluate the accuracy of our method and previous works on a fixed subset of 512 images randomly sampled from the test set for each type of corruption similar to Nie \textit{et al.}~\cite{DiffurePure}. 

\subsection{Experimental Results}

\textbf{CIFAR-10C}. Table.~\ref{tab:cifar10c-wideres} reports the experimental results against the 19 types of corruption on CIFAR-10C using WideResNet28-10 backbone. In CIFAR-10C, to make a fair comparison, we compared our method with DDA and the training-based methods since the rest of the diffusion model-based methods are sensitive to the hyperparameters following the previous works. Concretely, Compared with the state-of-the-art diffusion-based methods, our model improves the classification accuracy of all types of corruption. Concretely, PDDA improves by at least 3.18\% in contrast to corruption. Additionally, it achieves a significant improvement of 27.26\% on Gaussian noise corruption.
Meanwhile, the PDDA achieves 16.3\% average accuracy improvements, proving the proposed methods' validity. In the end, compared with the TENT, our method successfully reduces the gap between the training-based and diffusion-based methods, where our method even improves 1.3\% average accuracy. These results demonstrate that PDDA could alleviate the data shift and model shift to improve the accuracy of the classifier in TTA on the CIFAR-10C dataset.

\begin{table*}
    \centering
    \caption{Classification accuracy (\%) against the different types of corruption on CIFAR-100C under WideResNet-28-10, where Gaussian-N is the Gaussian noise.}
    \begin{adjustbox}{width={\textwidth},totalheight={\textheight},keepaspectratio}
    \setlength{\tabcolsep}{2px}{
    \begin{tabular}{c c c c c c c c c c c c c c c c c c c c c}  
        \toprule
        \textbf{Method} & \scriptsize  \textbf{Brightness} & \scriptsize \textbf{Contrast} & \scriptsize \textbf{Defocus} & \scriptsize \textbf{Elastic} & \scriptsize \textbf{Fog}& \scriptsize \textbf{Frost}& \scriptsize\textbf{Gaussian-B}& \scriptsize \textbf{Gaussian-N}&\scriptsize \textbf{Glass}& \scriptsize \textbf{Impulse}& \scriptsize \textbf{JPEG}& \scriptsize \textbf{Motion}& \scriptsize \textbf{Pixel}& \scriptsize \textbf{Saturate}& \scriptsize \textbf{Shot}&\scriptsize \textbf{Snow}&\scriptsize \textbf{Spatter}&\scriptsize \textbf{Speckle}&\scriptsize \textbf{Zoom}&\scriptsize \textbf{Avg.}  \\
        \midrule
        \textbf{Training-based}\\
        \midrule
        MEMO~\cite{MEMO} & \textbf{89.62} & \textbf{83.37}& \textbf{84.15}& \textbf{77.46} & \textbf{83.82}& \textbf{73.55}& \textbf{78.68}& \textbf{63.28}& \textbf{52.34}& 52.34& 57.48& \textbf{63.73}& 56.25& 60.83& 39.96& 41.18& 39.29& 31.36& 36.27& 61.31 \\
        TENT~\cite{Tent} & 74.22& 64.58& 72.92& 62.76& 66.15& 63.02& 62.24& 44.27& 49.48& 45.05& 47.4& 60.42& \textbf{60.68} & 58.85& 48.96& 55.21& 57.81& 49.74& \textbf{61.2}& 58.16 \\
        \midrule
        \textbf{Diffusion-based} \\
        \midrule
        DiffPure~\cite{DiffurePure} & 56.46& 38.13& 44.58& 40.42& 48.33& 34.79& 35.21& 21.04& 20.21& 23.54& 28.75& 37.08& 38.54& 53.33& 24.38& 40.83& 43.33& 25.0& 40.42& 36.55\\
        DDA~\cite{BacktoSource}& 40.2 & 31.86 & 46.86 & 41.37 & 29.22 & 32.35 & 43.43 & 40.29 & 42.84 & 38.92 & 43.82 & 43.73 & 43.04 & 45.39 & 44.41 & 37.94 & 41.76 & 39.31 & 44.22 & 40.58 \\
        PDDA (Our) & 71.56& 53.53& 64.02& 60.0& 63.43& 61.27& 55.49& 55.78& 41.78& \textbf{56.89}& \textbf{59.33}& 60.29& \textbf{55.2}& \textbf{62.84}& \textbf{65.11}& \textbf{59.12}& \textbf{64.31}& \textbf{63.33}& 61.11& \textbf{59.63} \\ 
        \bottomrule
    \end{tabular}}
    \end{adjustbox}
    \label{tab:cifar100c-wideres}
\end{table*}

\textbf{CIFAR-100C}. Similar to the CIFAR-10C, we compared our method with the training-based methods and DDA. Table.~\ref{tab:cifar100c-wideres} reports the experimental results against the 19 types of corruption on CIFAR-100C using WideResNet28-10 backbone. Compared with the DDA, we have achieved an average accuracy improvement of 16.99\%. It can be observed that we improved accuracy in all types of corruption compared with the DDA. This first demonstrates weaknesses in the diffusion-based method, specifically related to data shift and model shift. Then, it also proves that PDDA could mitigate label semantic loss based on the proposed three conditions.
Meanwhile, Compared with the training-based method, we reduce the gap of average accuracy from 17.58\% to 0.59\%. In cases of challenging corruption that result in significant semantic information loss, such as Speckle and Gaussian noise, PDDA improves accuracy by almost 11\%. This further confirms the effectiveness of PDDA.

\begin{table*}
    \centering
    \caption{Classification accuracy (\%) against the different types of corruption on ImageNet-C under ResNet-50, where Gaussian-N is the Gaussian noise. Training-based (N) means using N batch size to fine-tune the classifier.}
    \begin{adjustbox}{width={\textwidth},totalheight={\textheight},keepaspectratio}
    \setlength{\tabcolsep}{2px}{
    \begin{tabular}{c c c c c c c c c c c c c c c c c}
        \toprule
        \textbf{Method} & \scriptsize  \textbf{Brightness} & \scriptsize \textbf{Contrast} & \scriptsize \textbf{Defocus} & \scriptsize \textbf{Elastic} & \scriptsize \textbf{Fog}& \scriptsize \textbf{Frost}&  \scriptsize \textbf{Gaussian-N}&\scriptsize \textbf{Glass}& \scriptsize \textbf{Impulse}& \scriptsize \textbf{JPEG}& \scriptsize \textbf{Motion}& \scriptsize \textbf{Pixel}& \scriptsize \textbf{Shot}&\scriptsize \textbf{Snow}&\scriptsize \textbf{Zoom}&\scriptsize \textbf{Avg.}  \\
        \midrule
        \textbf{Training-based (N)}\\
        \midrule
        MEMO(N=64) &67.4& 12.3& 15.0& 47.5& 53.8& 47.5&  14.3& 15.1& 18.0& 34.3& 25.7 & 42.9 & 11.8& 41.5& 37.5&  32.3 \\ 
        MEMO(N=100) &\textbf{70.7}& 19.3& 16.0& \textbf{53.3}& 58.7& 42.0&  16.0& 20.7& 12.7& 38.7& 24.0& 42.0& 13.3& 43.3& 38.7& 34.0 \\ 
        TENT(N=64)& 69.0& 15.9& 17.0& 46.7& 53.6& 40.4&  15.4& 17.6& 17.6& 40.2& \textbf{34.8}& 43.5&  15.6& 47.3& \textbf{43.1}&34.5\\
        TENT(N=100)&70.3& 13.3& 13.5& 47.9& \textbf{61.7}& 48.2& 15.3& 17.5& 15.9& 35.2& 32.3& 43.2&  21.4& \textbf{45.3}& 42.7& 34.9 \\
        TENT(N=128)&70.1& 12.5& 16.1& 48.2& 58.5& \textbf{48.9}&  17.4& 20.1& 16.3& 37.1& 30.8& 45.1& 21.0& 42.6& 44.0 & 35.2\\ 
        \midrule
        \textbf{Diffusion-based} \\
        \midrule
        DiffPure~\cite{DiffurePure} & 58.2& 16.3& 18.4& 33.7& 26.5& 8.2& 5.1& 7.1& 11.2& 41.8& 14.3& 30.6& 7.1& 26.5& 26.5& 22.1\\
        DDA~\cite{BacktoSource}&60.9&21.1&11.9&27.2&44.6&28.3&28.3&18.0&17.0&47.6&22.1&33.4&32.3&24.2&28.3& 29.7 \\
        PDDA & 61.6& \textbf{23.2} & \textbf{23.2}& 21.2 & 41.4& 47.5& \textbf{41.4}&\textbf{22.2} &\textbf{49.5} &\textbf{49.5}& 24.2& \textbf{52.5}& \textbf{48.5}&34.3& 34.3&\textbf{38.3}\\ 
        \bottomrule
    \end{tabular}}
    \end{adjustbox}
    \label{tab:imagenet-resnet}
\end{table*}

\begin{table*}
    \centering
    \caption{Quantitative evaluations on ImageNet-C with different backbones. We compared PDDA with the latest diffusion model-based method to show the validity of our method. $\ast$ means it needs to fine-tune the classifier.}
    \begin{tabular}{c c c c c c c c}
    \toprule
    \textbf{Model} & \textbf{Source} & \textbf{MEMO} & \textbf{DiffPure} & \textbf{GDA} & \textbf{DDA} & \textbf{SDA}{$\ast$} &\textbf{PDDA} (Our) \\
    \midrule
         ResNet-50& 18.7 & 22.1 & 16.8 &  31.8 & 29.7 & 32.5 & \textbf{38.3}\\
         Swin-T& 33.1 & 29.5 & 24.8 &  42.2 & 40.0 & 42.5  & \textbf{43.9}\\
         ConvNext-T & 39.3 & 37.8 & 28.8 &  44.8 & 44.2 & 47.0 & \textbf{47.0}\\
    \bottomrule
    \end{tabular}
    \label{tab:all_results}
\end{table*}

\textbf{ImageNet-C}. We compared the latest diffusion-based adaptation, including DDA, GDA, and SDA, with PDDA, following the same settings of previous works. Concretely, Table~\ref{tab:imagenet-resnet} first reports the experimental results against the 15 types of corruption using the ResNet-50 backbone on 5-level severity corruption with the training-based methods. PDDA improves 4.3\% average accuracy compared to the MEMO using 100 batch size. Then, compared to the TENT, PDDA improves 3.1\% average accuracy. This shows the potential of diffusion-based adaptation. 

Then, Table~\ref{tab:all_results} compares the latest diffusion model-based methods with the different backbones, including DDA, GDA, and SDA. It can be noticed that PDDA improves 5.8\% and  1.4\% average accuracy in ResNet-50 and Swin-T, respectively, compared to the SDA, which achieves the best performance in all baselines. With the ConvNext-T backbone, PDDA improves 2.8\% compared to the GDA, the SOTA of the training-free diffusion-based adaptation methods. Compared to the SDA, PDDA achieves a close performance. Please note that SDA needs to re-train the classifier again and needs two pre-trained diffusion models. PDDA only needs a pre-trained unconditional diffusion model while being training-free. These results prove the effectiveness of PDDA and verify the validity of the two principles indicated in PDDA.

\textbf{ImageNet-W}. To further show the potential of the PDDA, we introduce an additional domain shift dataset, ImageNet-W. Then, we mainly compared DDA and SDA as the baseline for the PDDA in the ImageNet-W following the SDA setting. Concretely, Table~\ref{tab:all_results_imagenet-w} shows that PDDA achieves 0.7\% and 1.8\% average accuracy using the ResNet-50 and Swin-T classifier compared to the SDA. With the ConvNext-T backbone, compared to the SDA, the training required diffusion-based adaptation, and PDDA achieves a close performance while only having a 0.1\% average accuracy gap. This proves the validity and efficiency of PDDA.

\begin{table*}
    \centering
    \caption{Quantitative evaluations on ImageNet-W with different backbones. We compared PDDA with the latest diffusion model-based method to show the validity of our method. $\ast$ means it needs to fine-tune the classifier.}
    \begin{tabular}{c c c c c c}
    \toprule
    \textbf{Model} & \textbf{Source} &  \textbf{DiffPure} & \textbf{DDA} & \textbf{SDA}{$\ast$} &\textbf{PDDA} \\
    \midrule
         ResNet-50& 37.7 & 29.1 & 52.8 & 54.7 & \textbf{55.4} \\
         Swin-T&  66.5 &  52.7 & 65.9 &  67.3 & \textbf{69.1}\\
         ConvNext-T & 67.6 &  55.8 & 67.9 &  \textbf{69.4} &69.3\\
    \bottomrule
    \end{tabular}
    \label{tab:all_results_imagenet-w} 
\end{table*}

To sum up, we report the experimental results of PDDA among Four datasets. Compared with the diffusion-based adaptation. PDDA has improved CIFAR-10C, CIFAR-100C, ImageNet-C, and ImageNet-W. Meanwhile, we report the experimental results compared with the latest training-based methods to prove the effectiveness further. According to the results, PDDA could achieve close performance and even make a few improvements partly due to corruption, especially for noise type corruption, including Gaussian and impulse noise. In the end, PDDA is based on the DPM-Solver++, which maintains a low time cost.

\subsection{Ablation Study} 
\begin{table}
    \centering
    \caption{Ablation study for combinations of each component in PDDA based on ImageNet-C with ResNet-50, where Sam is the sampling strategy.}
    \begin{tabular}{c c c | c c}
    \toprule
    \textbf{$f^1$} & \textbf{$f^2$} & \textbf{Sam} &\textbf{Avg. (\%)} \\
    \midrule
     $\surd$ & & $\surd$ & 17.51  \\
     & $\surd$ & $\surd$ & 22.56 \\
      & &  $\surd$ & 10.89  \\
    $\surd$ & $\surd$ & & 28.03   \\
    $\surd$ & $\surd$ & $\surd$ & \textbf{38.3}   \\
    \bottomrule
    \end{tabular}
    \label{tab:alabation study}
\end{table}
To show the effectiveness of our method, we report the ablation study for the combinations of the guidance. We also report more ablation studies, including the influence of the ensemble strategy and hyperparameter $t^{\ast}$ for $f^1$. 

\begin{table*}
    \centering
    \caption{ Ablation study for patch-aware method against the different types of corruption on ImageNet-C under ResNet-50, where Gaussian-N is the Gaussian noise. \textbf{w/} means using the patch-aware method and \textbf{w/o} means using all features to calculate $f^{1}$ each time.}
    \begin{adjustbox}{width={\textwidth},totalheight={\textheight},keepaspectratio}
    \setlength{\tabcolsep}{2px}{
    \begin{tabular}{c c c c c c c c c c c c c c c c c}
        \toprule
        \textbf{Method} & \scriptsize  \textbf{Brightness} & \scriptsize \textbf{Contrast} & \scriptsize \textbf{Defocus} & \scriptsize \textbf{Elastic} & \scriptsize \textbf{Fog}& \scriptsize \textbf{Frost}&  \scriptsize \textbf{Gaussian-N}&\scriptsize \textbf{Glass}& \scriptsize \textbf{Impulse}& \scriptsize \textbf{JPEG}& \scriptsize \textbf{Motion}& \scriptsize \textbf{Pixel}& \scriptsize \textbf{Shot}&\scriptsize \textbf{Snow}&\scriptsize \textbf{Zoom}&\scriptsize \textbf{Avg.}  \\
        \midrule
        PDDA (w/o)  & 61.0& \textbf{25.0} & \textbf{24.0}& \textbf{22.0} & \textbf{43.0}& 38.0& \textbf{44.0}& 19.0& 47.0& 48.0& 24.0& 51.0& 46.0&\textbf{36.0}& \textbf{37.0}&37.7\\ 
        PDDA (w/)  & \textbf{61.6}& 23.2 & 23.2& 21.2 & 41.4& \textbf{47.5}& 41.4&\textbf{22.2} &\textbf{49.5} &\textbf{49.5}& \textbf{24.2}& \textbf{52.5}& \textbf{48.5}&34.3& 34.3&\textbf{38.3}\\
        \bottomrule
    \end{tabular}}
    \end{adjustbox}
    \label{tab:feature}
\end{table*}
\textbf{Combinations.} To prove the effectiveness of the proposed method, we report the ablation study for the combination of the guidance reported in Table~\ref{tab:alabation study}. It can be shown that using each condition alone will decrease the performance of the PDDA. This further shows the validity of the two principles since lacking one will significantly decrease the performance of the PDDA. 

\begin{table*}
    \centering
    \caption{ Ablation study for gradient projection method against the different types of corruption on ImageNet-C under ResNet-50, where Gaussian-N is the Gaussian noise. \textbf{w/} means using gradient projection method and \textbf{w/o} means not using gradient projection method.}
    \begin{adjustbox}{width={\textwidth},totalheight={\textheight},keepaspectratio}
    \setlength{\tabcolsep}{2px}{
    \begin{tabular}{c c c c c c c c c c c c c c c c c}
        \toprule
        \textbf{Method} & \scriptsize  \textbf{Brightness} & \scriptsize \textbf{Contrast} & \scriptsize \textbf{Defocus} & \scriptsize \textbf{Elastic} & \scriptsize \textbf{Fog}& \scriptsize \textbf{Frost}&  \scriptsize \textbf{Gaussian-N}&\scriptsize \textbf{Glass}& \scriptsize \textbf{Impulse}& \scriptsize \textbf{JPEG}& \scriptsize \textbf{Motion}& \scriptsize \textbf{Pixel}& \scriptsize \textbf{Shot}&\scriptsize \textbf{Snow}&\scriptsize \textbf{Zoom}&\scriptsize \textbf{Avg.}  \\
        \midrule
        PDDA (w/o)  & 55.6& \textbf{27.3} & 21.2& \textbf{37.3} & 37.3& 30.3& 24.2& \textbf{23.2}& 24.2& 48.5& 20.2& 41.5& 30.3&30.3& 27.3&31.9\\ 
        PDDA (w/)  & \textbf{61.6}&23.2 & \textbf{23.2}& 21.2 & \textbf{41.4}& \textbf{47.5}& \textbf{41.4}&22.2 &\textbf{49.5} &\textbf{49.5}& \textbf{24.2}& \textbf{52.5}& \textbf{48.5}&\textbf{34.3}& \textbf{34.3}&\textbf{38.3}\\
        \bottomrule
    \end{tabular}}
    \end{adjustbox}
    \label{tab:gradient}
\end{table*}
\textbf{Patch-aware method.} To show the effectiveness of the proposed patch-aware method in the semantic keeper, we show the ablation study with and without it shown in Table~\ref{tab:feature}. To make a fair comparison, we use all the features extracted from all layers of the UNet to replace the patch-aware method and calculate the constrative loss individually. It can be shown that the patch-aware method improved 0.6\% average accuracy and only needs to calculate the constrative loss once, which verifies its validity.

\textbf{Gradient projection method.} To show the influence of the conflict between the $f^1$ and $f^2$, we show the ablation study with and without using the gradient projection in Table~\ref{tab:gradient}. It can be noticed that the gradient projection method improves average accuracy from 31.9\% to 38.3\%, which illustrates there is a conflict and further verifies the validity of our method.

\subsection{Visualization}
To show the effectiveness of PDDA, we report the additional visualization results.

\begin{figure*}[h!]
    \centering
    \includegraphics[width=\textwidth]{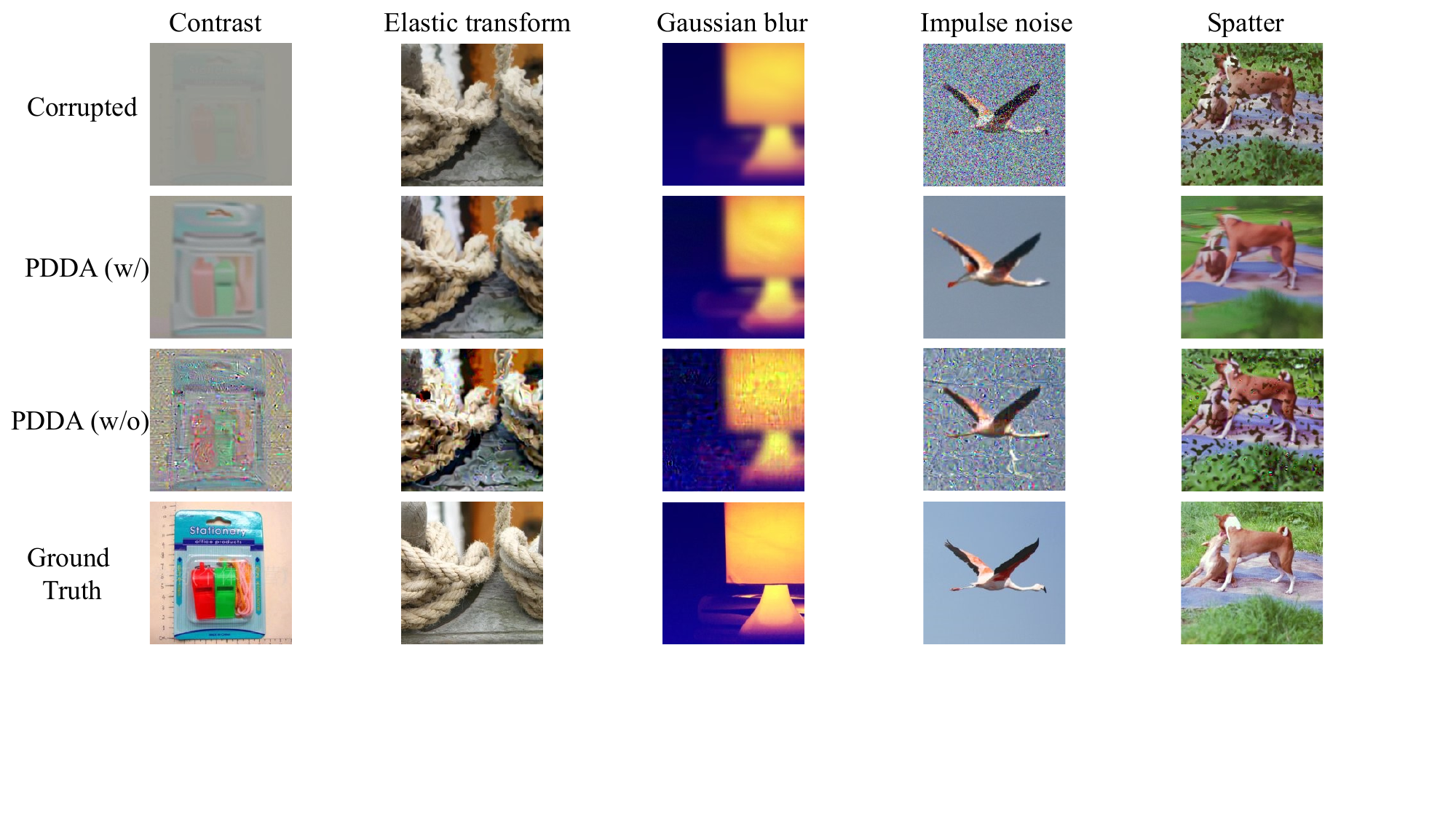}
    \caption{Visualization of ablation study for sampling strategy against different corruption, where PDDA (w/) means we use the sampling strategy and PDDA (w/o) means we do not use the proposed sampling strategy, and ground truth is the image in the training domain, i.e., without corruption.} 
    \label{fig:differentmodels}
\end{figure*}

Fig.~\ref{fig:differentmodels} reports the visualization of PDDA compared with the baselines. It was found that PDDA maintains semantics as much as possible. In contrast to corruption, PDDA can recover some of the semantics. Meanwhile, PDDA can remove the extra noise and keep the label semantic in spatter corruption, proving its effectiveness. Meanwhile, we also report the additional ablation study about the sampling strategy to illustrate the improvement. It could be found that the sampling strategy could improve the quality of the generated image, proving the sampling strategy's necessity. Meanwhile, it also shows that the sampling strategy is important for the acceleration method in diffusion-based methods, which is an additional finding in our work.

\begin{figure*}[h!]
    \centering
    \includegraphics[width=\textwidth]{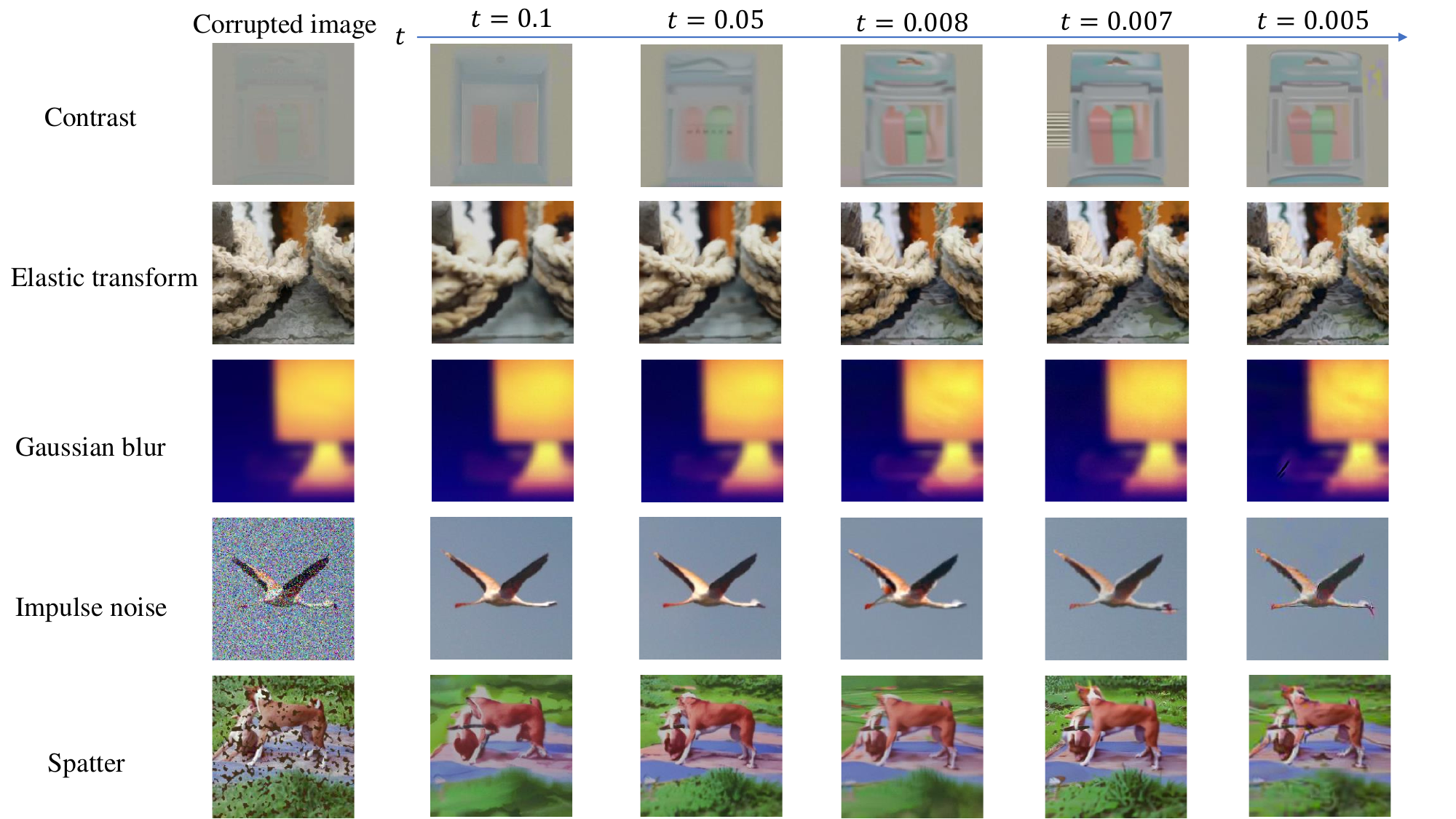}
    \caption{Visualization of ablation study for $t^{\ast}$ in $f^1$ against different corruption.} 
    \label{fig:abationstudy_time}
\end{figure*}

\textbf{Hyperparameter.} PDDA contains one hyperparameter: the special time $t^{\ast}$ for $f^1$. We report the influence of the hyperparameter on the quality of the generated images shown in Fig.~\ref{fig:abationstudy_time}. Previous works~\cite{Diffuse} show that the diffusion model will focus on the local details when the $t$ tends to be $0$. However, the corruption's extra noise, such as the impulse noise, will influence the generated images if we directly set $t=0$. In this condition, we slightly increase the $t$ to let the diffusion model focus on the local details and ignore part of the noise. As shown in Fig.~\ref{fig:abationstudy_time}, when $t=0.1$, the diffusion model starts to drop the local detail since it can be found that the quality of the generated image against contrast is decreasing. In internal $t\in[0.05,0.008]$, the generated images are high quality compared with the interval $t\in[0.007,0.005]$. Therefore, choosing the value in internal $[0.05,0.008]$ can get a great result. In our work, we directly set $t^{\ast}=0.008$.

\section{Conclusion}
We proposed the PDDA with two principles, i.e., preserving the semantics (P1) with as minor modification as possible (P2) for diffusion-based TTA. Motivated by the two principles, we propose a semantic keeper and a modification keeper to achieve the diffusion-based method for TTA without the complex design. The experimental results demonstrate that PDDA gets state-of-the-art results. Compared with the latest diffusion-based methods, our method gets a better result for some types of corruption and even the average accuracy in CIFAR-10C, CIFAR-100C, ImageNet-W, and ImageNet-C with different backbones. Meanwhile, for the difficult ImageNet-C with 5-level severity, we achieve state-of-the-art performance based on the DPM-Solver++.

We also report the other findings in our work: 1) The acceleration method is sensitive to the gradient of guidance, which makes it difficult to implement the guidance method directly on the entire reverse process. 2) The diffusion-based methods are ineffective for blurring type corruptions mentioned in the supplementary material. We will further explore these conditions.

\clearpage
\bibliographystyle{IEEEtran}
\bibliography{reference}
\begin{IEEEbiography}[{\includegraphics[width=1in,height=1.25in,clip,keepaspectratio]{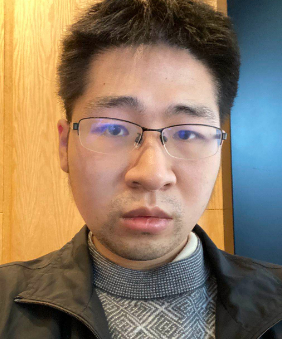}}]{Kaiyu Song}
	received his B.S. and M.S. from Chongqing University and Yunnan University in 2018 and 2021, respectively. He is currently pursuing a Ph.D degree at Sun Yat-Sen University, where his advisor is Prof.HanJiang Lai. His research interests include computer vision, deep learning, and diffusion models. He was the reviewer of UAI, ACM MM, NeurIPS, and ICLR.
\end{IEEEbiography}

\begin{IEEEbiography}[{\includegraphics[width=1in,height=1.25in,clip,keepaspectratio]{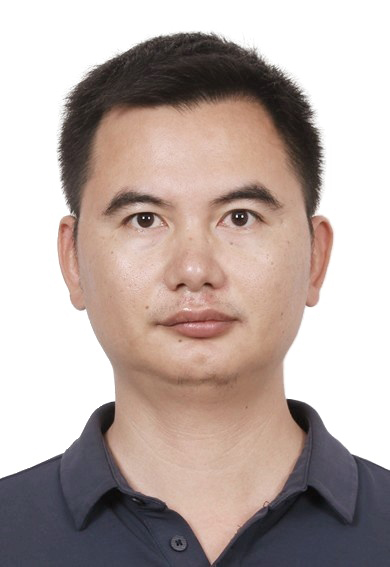}}]{Hanjiang Lai}
	received his B.S. and Ph.D. degrees from Sun Yat-sen University in 2009 and 2014, respectively. He worked as a research fellow at the National University of Singapore during 2014-2015. He is currently an associate professor at Sun Yat-Sen University. His research interests include machine learning algorithms, deep learning, and computer vision.
\end{IEEEbiography}

\begin{IEEEbiography}[{\includegraphics[width=1in,height=1.25in,clip,keepaspectratio]{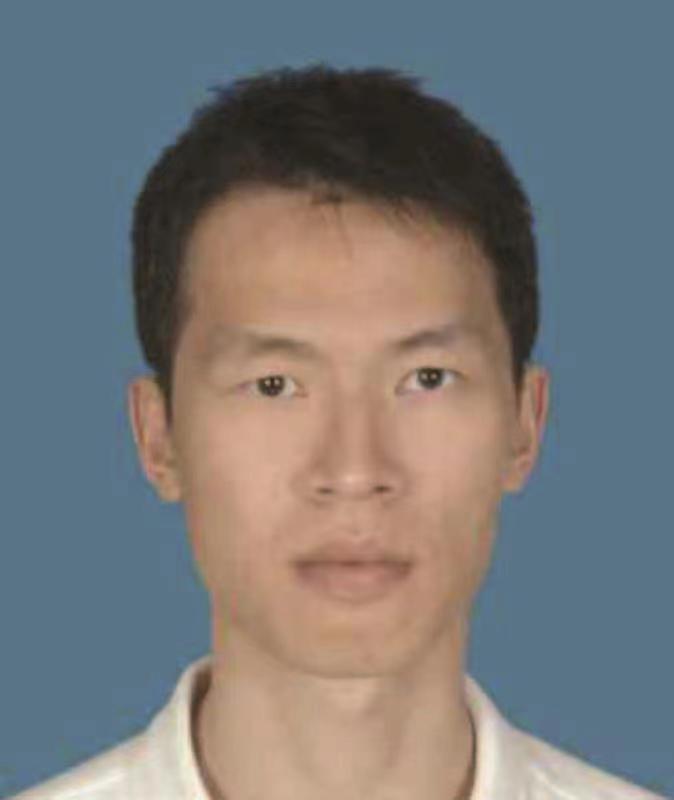}}]{Yan Pan}
 received a B.S. degree in information science and a Ph.D. degree in computer science from Sun Yat-Sen University, Guangzhou, China, in 2002 and 2007, respectively. He is currently an Associate Professor at Sun Yat-Sen University. His current research interests include hashing methods, image retrieval, and causal inference. 
\end{IEEEbiography}

\begin{IEEEbiography}[{\includegraphics[width=1in,height=1.25in,clip,keepaspectratio]{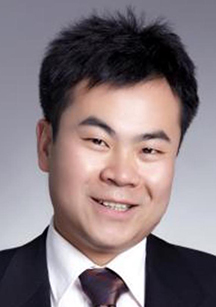}}]{Kun Yue}
 received his M.S. and Ph.D. in computer science from Fudan University in 2004 and Yunnan University in 2009, respectively. He is currently a professor at Yunnan University and the dean of Yunnan Key Laboratory of Intelligent Systems and Computing. He has authored over 80 papers in peer-reviewed journals and conferences, such as IEEE TCYB, IEEE TSC, DMKD, AAAI, CIKM, and UAI. His current research interests include knowledge engineering and Bayesian deep learning.
\end{IEEEbiography}

\begin{IEEEbiography}[{\includegraphics[width=1in,height=1.25in,clip,keepaspectratio]{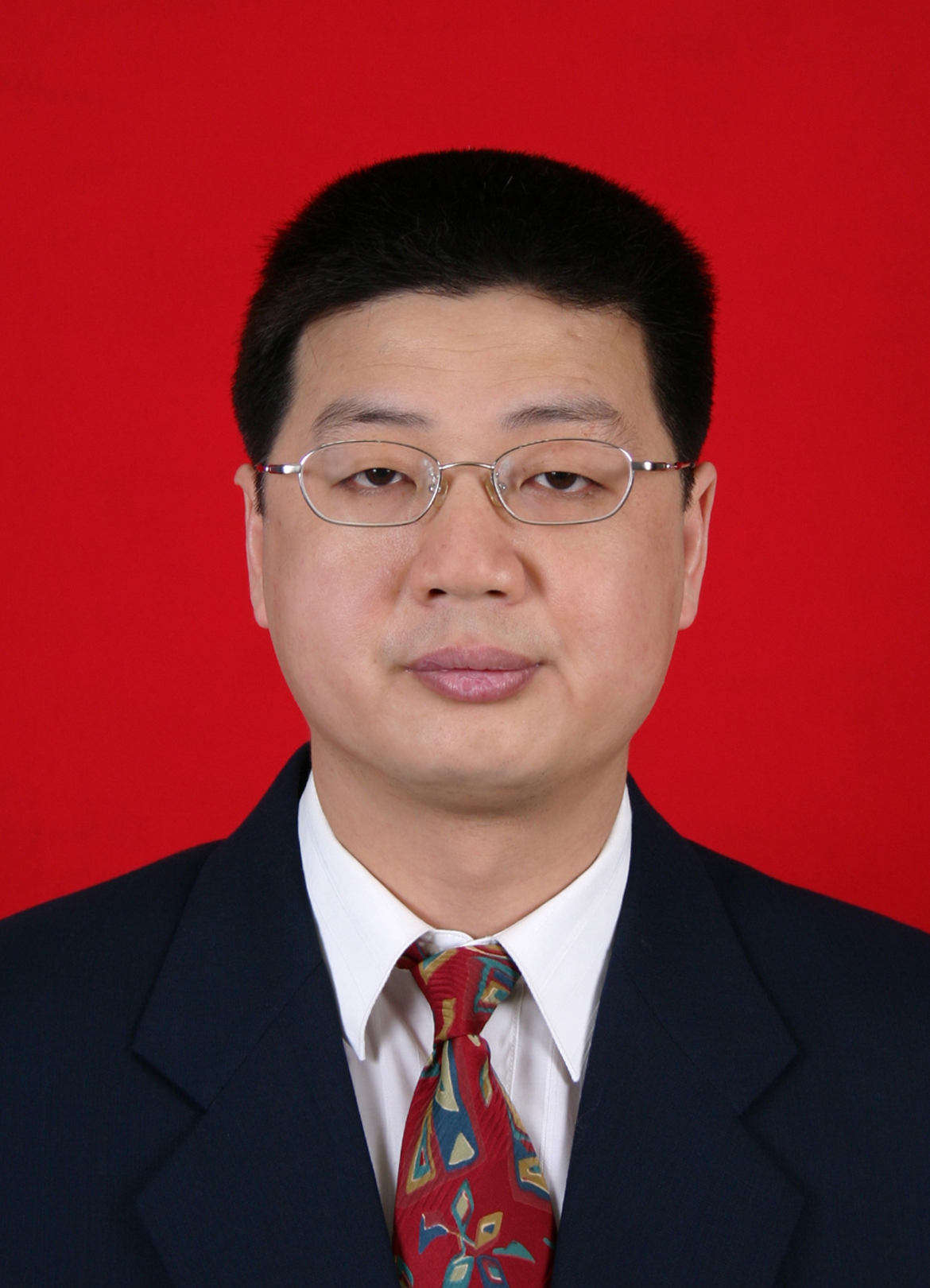}}]{Jian Yin}
 received the B.S., M.S., and Ph.D. degrees from Wuhan University, China, in 1989, 1991, and 1994, respectively, all in computer science. He joined Sun Yat-Sen University in July 1994, and now he is a professor at the School of Artificial Intelligence. He has published more than 200 refereed journal and conference papers. His current research interests are in the areas of artificial intelligence, and machine learning. He is a senior member of the China Computer Federation.
\end{IEEEbiography}

\end{document}